# Development of a CAV-based Intersection Control System and Corridor Level Impact Assessment

**Ardeshir Mirbakhsh**
Department of Transportation Engineering
New Jersey Institute of Technology, Newark, New Jersey, 07102
Email: am2775@njit.edu

**Joyoung Lee**
Department of Transportation Engineering
New Jersey Institute of Technology, Newark, New Jersey, 07102
Email: jo.y.lee@njit.edu

**Dejan Besenski**
Department of Transportation Engineering
New Jersey Institute of Technology, Newark, New Jersey, 07102
Email: besenski@njit.edu

Word Count: 6707 words + 2 table (250 words per table) = 7,207 words

*Submitted 07/30/2022*

*Ardeshir Mirbakhsh, Joyoung Lee, and Dejan Besenski*

**ABSTRACT**

This paper presents a signal-free intersection control system for CAVs by combination of a pixel reservation algorithm and a Deep Reinforcement Learning (DRL) decision-making logic, followed by a corridor-level impact assessment of the proposed model. The pixel reservation algorithm detects potential colliding maneuvers and the DRL logic optimizes vehicles' movements to avoid collision and minimize the overall delay at the intersection. The proposed control system is called Decentralized Sparse Coordination System (DSCLS) since each vehicle has its own control logic and interacts with other vehicles in coordinated states only. Due to the chain impact of taking random actions in the DRL's training course, the trained model can deal with unprecedented volume conditions, which poses the main challenge in intersection management. The performance of the developed model is compared with conventional and CAV-based control systems, including fixed traffic lights, actuated traffic lights, and the Longest Queue First (LQF) control system under three volume regimes in a corridor of four intersections in VISSIM software. The simulation result revealed that the proposed model reduces delay by 50%, 29%, and 23% in moderate, high, and extreme volume regimes compared to the other CAV-based control system. Improvements in travel time, fuel consumption, emission, and Surrogate Safety Measures (SSM) are also noticeable.

**INTRODUCTION**

Due to rapid population growth and the increased number of vehicles, traffic congestion, collisions, and pollution have become leading causes of decreased living standards. In only 2017 and the USA, traffic congestion has caused 8.8 billion hours of delay and 3.3 billion gallons of fuel waste, resulting in a total cost of 179 billion USD. Several studies prove that human errors are pivotal in traffic congestion and accidents, and driver error contributes to up to 75% of roadway crashes worldwide. Intersections are crucial in traffic delays and collisions among urban traffic facilities. In the United States, 44% of reported traffic accidents occur at urban intersections, leading to 8,500 fatalities and around 1 million injuries annually (3).

Taking advantage of the communication capability of Connected and Autonomous Vehicles (CAV) to improve traffic measures and safety at intersections is attracting noticeable attention from the transportation research world. Several approaches, such as online optimization, trajectory planning, and rule-based intersection control logic for CAVs, have been around for the last couple of decades. Although a comprehensive literature review study on CAV-based intersection management systems shows that Machine Learning (ML)-based approaches are more promising than other methodologies, less than 4% of the proposed systems from 2008 to 2019 are ML-based (2). And most of them have used ML to optimize traffic lights' phasing and timing in an isolated or grouped setting.

Among ML techniques, Deep Reinforcement Learning (DRL) is the closest to the human brain's learning system, outperforming humans in solving complicated decision-making problems, such as strategic multiplayer video games. Therefore it appears to be a well-fitting solution to tackle stochastics in CAV's control at intersections (4) (5). Few pieces of literature exist about DRL-based intersection control systems, and no DRL-based intersection control has been developed to control individual vehicles' movement in roadway circumstances.

In this study, a Decentralized Sparse Coordination Learning System (DSCLS) controls CAVs at intersections. Each vehicle approaching the intersection is a separate agent and coordinates with other agents only if required. The intersection is divided into grid areas (known as cells), each agent tries to reserve its desired cells ahead of time to detect potential colliding movements. A DRL coordination logic will control the vehicles to avoid collisions and minimize overall delay at the intersection. The DSCLS is developed as a Python application controlling individual vehicles in VISSIM software through the Component Object Model (COM) interface. Performance of the proposed model is compared with other conventional and CAV-based intersection control systems in a VISSIM test bed.

**CAV-BASED INTERSECTION CONTROL SYSTEMS**

Based on the existing literature, CAV-based intersection control systems are clustered into four main groups, including 1) rule-based, 2) online optimization-based, 3) trajectory planning-based, and 4) ML-based algorithms.

**Rule-based Intersection Control Systems for CAVs**

Most rule-based intersection logics are based on the intersection space reservation approach. The space reservation approach was initially developed in 2004 by Dresner et al. (6). This method divides the intersection area into an n×n grid of reservation tiles. Each approaching vehicle to the intersection attempts to reserve a time-space block at the intersection area by transmitting a reservation request to the intersection manager, including its information such as speed and location. According to intersection control policy, the intersection manager decides whether to approve and provide more passing restrictions to the vehicle or reject the reservation request. Dresner et al. deployed a "First Come First Serve" (FCFS) control policy, in which the passing priority is assigned to the vehicle with the earliest arrival time. All other vehicles must yield to the vehicle with priority. In a following study, Dresner et al. (7) added several complementary regulations to FCFC policy to make it work more reliable, safe, and efficient. Simulation results revealed that FCFS policy noticeably reduces intersection delay compared to traffic light and stop sign control systems.

Zhang et al. (8) proposed a state-action control logic based on Prior First in First Out (PriorFIFO) logic. They assumed spatial-temporal and kinetic parameters for vehicle movement based on a centralized scheduling mechanism. This study aimed to reduce control delay for vehicles with higher priority. The simulation results with a combination of high, average, and low priority vehicles showed that the algorithm works well for vehicles with higher priority. Meanwhile, causing some extra delays for regular vehicles with lower priority.

Carlino et al. (9) developed an auction-based intersection control logic based on Clarke Groves tax mechanism and pixel reservation. In this approach, if commonly reserved tiles exist between vehicles, an auction is held between the involved vehicles. All vehicles in each direction contribute to their leading vehicle to win the auction, and the control logic decides which leading vehicles receive a pass order first. The bid's winner and its contributors (followers) have to pay the runner-up bid amount with a proportional payment (based on their contribution value in the bid). A "system wallet" component was added to auction-based intersection control to ensure low-budget vehicles or emergency vehicles would not be over-delayed. A comparison of simulation results showed that the auction-based control logic outperforms the FIFO logic.

**Online Optimization Intersection Control logic for CAVs**

Yan et al. (10) proposed a dynamic programming-based optimization system to find optimal vehicle passing sequence and minimize intersection evacuation time. In this algorithm, the optimizer agent clusters vehicles into several groups so that each group of vehicles can pass the intersection simultaneously without a potential collision. Since vehicles were clustered into several groups, the conventional problem of finding an optimal passing order of vehicles was transformed into partitioning the vehicles into different groups and finding optimal group sequences to minimize the vehicle evacuation time. In this method, approaching vehicles at the intersection provide their data to the controller agent. And the controller agent has to run the dynamic programming optimization whenever a new vehicle is detected. However, if a group of vehicles is authorized to pass the intersection, the calculation process is delayed till the whole group passes the intersection.

Wu et al. (11) deployed a timed Petri Net model to control a simple intersection with two conflicting movements. The intersection control was considered a distributed system, with parameters such as vehicles' crossing time and time-space between successive vehicles. Two control logics, including 1) central controller and 2) car-to-car communication, were developed and tested. In central controller logic, approaching vehicles to the intersection provides their information to the controller center, and the controller center has to provide optimized passing order for the vehicles. While in car-to-car control logic, each leading vehicle collects its followers' data, and if the distance between them is shorter than a threshold, they would be considered a group. In the latter system, only leading vehicles of groups communicate with each other, and the passing order priority for each group is defined by its leader's proximity to the intersection. The optimization task was decomposed into chained sub-problems by a backtracking process, and the final optimal solution was found by solving single problems and applying forward dynamic programming to find the shortest path from the original problem to the last problem in the graph. The simulation results revealed that both control logics have the same performance in delay reduction, resulting in almost the same queue length.

Fayazi et al. (12) deployed Mixed-Integer Linear Programming (MILP) approach to optimize CAVs' arrival time at the intersection. In this approach, the intersection controller receives arrival and departure times from approaching vehicles and optimizes the vehicles' arrival times. The optimization goal was to minimize the difference between the current time and the last vehicle's expected arrival time at the intersection. To ensure all vehicles are not forced to travel near the speed limit, a cost value was defined as a function of the difference between the assigned and desired crossing time for each vehicle. Several constraints, such as speed limit, maximum acceleration, minimum headway, and minimum cushion for conflicting movements, were applied to the model. A two-movement intersection simulation results showed that the MILP-based controller reduces average travel time by 70.5% and average stop delay by 52.4%. It was also proved that the control logic would encourage platooning under a specific gap setting.

Lee et al. (13) proposed a Cumulative Travel-time Responsive (CTR) intersection control algorithm under CAVs' imperfect market penetration rate. They considered the elapsed time spent by vehicles from when they entered the network to the current position as a real-time measure of travel time. Kalman filtering approach was deployed to cover the imperfect market penetration rate of CAVs for travel time estimation. Simulations were run in VISSIM for an isolated intersection with 40 volume scenarios covering the volume capacity ratio ranging from 0.3 to 1.1 and different CAV market penetration rates. Simulation results showed that the CTR algorithm improves mobility measures such as travel time, average speed, and throughput by 34%, 36%, and 4%, compared to the actuated control system. The $CO_2$ emission and fuel consumption were also reduced by 13% and 10%. It was also revealed that the CTR would produce more significant benefits as the market penetration rate passes the threshold of 30%. In general, more benefits were observed as the total intersection volume increased.

**Trajectory Planning Intersection Control Logics for CAVs**

Lee et al. (14) proposed a trajectory planning-based intersection control logic named Cooperative Vehicle Intersection Control (CVIC). The control algorithm provided a location-time diagram (trajectory) of individual vehicles and minimized the length of overlapped trajectories (conflicting vehicles). The optimization task was modeled as a Nonlinear Constrained Programming (NCP) problem. To ensure an optimal solution is achieved, three analytical optimization approaches, including the Active Set Method (ASM), Interior Point Method (IPM), and Genetic Algorithm (GA), were deployed. All three algorithms were implemented in parallel, and the first acceptable solution was implemented. Minimum/Maximum acceleration, speed, and safe headway were considered optimization problem's constraints. A simulation testbed was developed in VISSIM to compare the proposed algorithm's efficiency with actuated traffic signals under different volume conditions. The simulation results revealed that the proposed algorithm improves stopped delay, travel time, and total throughput by 99%, 33%, and 8%. A 44% reduction in $CO_2$ emission and fuel consumption was also achieved. It was also revealed that the CVIC is more advantageous when the intersection operates under oversaturated conditions.

Based on the assumption that vehicle trajectories can be defined as cubic interpolated splines, having the flexibility to reflect delays from the given signal timing, Gutesta et al. (15) developed a trajectory-driven intersection control for the CAVs. Vehicle trajectories were developed for several vehicles passing multiple intersections. The optimization goal was to minimize the sum of all trajectory curves (which reflect control delays) conditioned to meeting safety constraints. Traffic constraints such as speed limits were reflected in the model by adjusting the trajectory slopes. A Genetic Algorithm (GA) was deployed to evaluate all possible combinations of optimized single-vehicle trajectories. In addition, an Artificial Neural Network (ANN), trained with available traffic stream factors, was deployed to achieve short-term prediction of vehicle delays to be integrated with the optimization model. The simulation results revealed that the proposed control algorithm improves traffic measures under stable and unstable traffic conditions, even with CAVs' low market penetration rate.

Krajewski et al. (16) proposed a decoupled cooperative trajectory optimization logic to optimize and coordinate CAVs trajectories at signalized intersections. The optimization goal was to minimize delay by coordinating conflicting movements such as straight going and left turns. The state-space of each vehicle had three dimensions: position, speed, and time. The optimization task was transformed into a graph (nodes representing the states and edges representing possible transitions between pairs of states) with a combined cost function of delay and comfort. The "decoupled" term refers to splitting the optimization problem into two stacked layers, 1) Trajectory Layer (TL) and 2) Negotiation Layer (NL). The TL layer's task was to calculate trajectories for individual vehicles, and the NL layer's task was to coordinate all trajectories to prevent potential collisions. The latter task was achieved by setting constraints for each vehicle's TL algorithm, and the cost function was a weighted sum of the individual cost functions. The weight could also prioritize specific vehicle types. Simulation results revealed that the proposed algorithm reduces the cost value by 28% compared to the intelligent driver model replicating human-driven vehicles.

**Machine Learning-Based Intersection Control Logics for CAVs**

Recent access to abundant and cheap computation and storage resources has made ML approaches popular in solving stochastic problems in different fields. Over the last few years, some researchers have deployed ML techniques to optimize single/multi-intersection traffic signal controls or develop signal-free control logic for CAVs. A selection of ML-based interaction control systems is reviewed in this section.

Lamouik et al. (18) developed a multiagent control system based on Deep Reinforcement Learning to coordinate CAVs at the intersection. Each vehicle transfers five features to the controller agent in this method, including position, speed, dimension, destination, and priority. The controller agent has three possible responses for each vehicle, including acceleration, deceleration, and keep-same-speed. Reward value is a function of speed, priority, and collision. The controller agent was expected to avoid collisions and prioritize vehicles with higher priority or higher speed in the training process. The controller agent was trained after several training epochs. However, the simulations were not run in a traffic network setting, and no traffic measures were assessed in this study.

Tong Wu et al. (19) developed a multiagent deep reinforcement learning algorithm to optimize several traffic lights in a corridor. Each traffic light was considered an agent, with actions being green lights for different phases. The reward value was set as a function of real-time delay for each vehicle, weighted by its priority. The innovation in this study was information exchange between controller agents so that each agent could estimate the policy of the other agents. According to other agents' estimated policy, each agent could adjust the local policy to achieve the optimal global policy. Two networks with different intersections numbers were developed in SUMO for assessment purposes. Simulation results revealed that the proposed model outperforms independent deep Q-learning (no exchange of information between agents) and deep deterministic policy gradient (single-agent controlling all intersections). It was also shown that the model outperforms self-organizing and fixed-time traffic lights.

Touhbi et al. (20) developed a Reinforcement Learning (RL)-based adaptive traffic control system. The main goal was to find the impact of using different reward functions, including queue length, cumulative delay, and throughput, on intersection performance. Unlike previous studies, which considered queue or delay as states, in this study, the state was defined as a maximum residual queue (queue length divided by lane length) to reflect the traffic load on each phase. The learning process for a four-way single-lane intersection took 100 epochs of 1hr simulation runs, and the simulation results revealed that the proposed algorithm remarkably outperforms the pre-timed traffic signals. The analysis of different reward function deployments showed that each function's efficiency highly depends on the traffic volume.

Yuanyuan Wu et al. (21) developed a decentralized coordination algorithm for CAVs' management at the intersection. Each vehicle was considered an agent, and each agent's state included its lane, speed, and moving intention. The intersection area was divided into n×n grid, and each vehicle was supposed to reserve its desired pixel ahead of time. Vehicles could enter either a coordinated or an independent state, based on having a reserved pixel in common or not. A Conventional multiagent Q-learning was deployed in this study, meaning that new states and their Q-values were added to a matrix as they came up, and the controller agent had to refer to a specific cell in the matrix to make a decision. Aligned with the multiagent optimization goal, each agent's effort to maximize its Q-value resulted in a maximized global reward for all agents. Simulation results revealed that the proposed model outperforms FCFS, fixed time traffic light, and LQF control system in delay reduction.

## METHODOLOGY

### Reinforcement Learning and Markov Decision Process

RL problems in which an agent learns to solve a problem are a subfield of artificial intelligence and Markov Decision Processes (MDPs), first proposed in the 1950s (4). The process can be expressed as an agent taking actions; the environment responding to the agent by presenting new state and reward based on the last act's quality. The agent seeks to maximize the reward over a series of interactions with the environment in discrete time steps (t). The MDP framework appears in **Figure 1**.

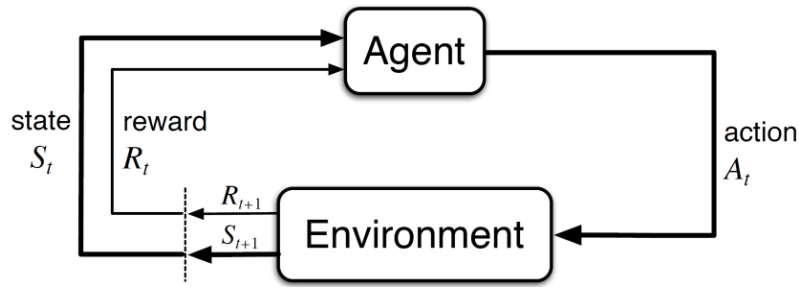

**Figure 1 The agent-environment interactions in the MDP (4)**

The main idea of RL is to run an experiment several times (known as an epoch or episode) to train an agent to take the best actions. The training course consists of running several iterations of the experiment until the cumulative reward for each epoch converges to a maximum value. Each epoch ends with losing or winning an Atari game player agent. However, for the experiment with nonbinary outcomes, such as automated intersection control systems, each epoch could be running the experiment for a fixed period.

### Deep Reinforcement Learning and Deep Q-Networks

In conventional RL, the return values for possible actions in each state are saved in a multi-dimensional matrix. However, this approach is not practical in complex environments since a significant space is required to save all state-action combinations. Moreover, the agent will never know how to deal with an unprecedented state.

Meanwhile, in DRL, the look-up tables are substituted with neural networks, eliminating the shortcomings of conventional RL. Several methods, including Reinforce, State-Action-Reward-State-Action (SARSA), and Deep Q-Network (DQN), are proposed to perform the DRL process. In this study, DQN is used, which was initially proposed by Mnih et al. (22). The DQN estimates every action's expected return using Temporal Difference learning. DQN formulation for estimating return values, called Q-function, is presented in Equation (1).

$$Q(s, a) = r + \Upsilon \max Q(s', a') \qquad (1)$$

Where:

- $Q(s, a)$ or Q-value defines the value of the current state-action pair.
- $Q(s', a')$ defines the value of the next state-action pair.

- Ɣ or "discount rate" ranges from zero to one and defines the present value of future rewards. If Ɣ = 0, the agent is myopic, only concerned with maximizing the immediate reward. As Ɣ approaches one, the agent cares more about future rewards and becomes more farsighted.
- $r$ is the immediate reward for the current actions.

AI training and trial and error processes are usually performed in simulation testbeds, enabling a trained agent to tackle a broad scope of real-world scenarios. In the DQN approach, states are the input layer of the neural network, and the neural network's output layer is an estimation of Q-values for all possible actions in the current state. The training occurs by updating the neural network weights based on a batch of recent historical data ((states, actions), Q-values). DQN is an off-policy algorithm, meaning no policy is used to generate training data for DQN, and the algorithm learns from all experiences gathered by an agent. In this study, the simulation test bed is developed in VISSIM software, and the DQN is developed as a Python application; a high-level training process of a DQN agent appears in **Figure 2**.

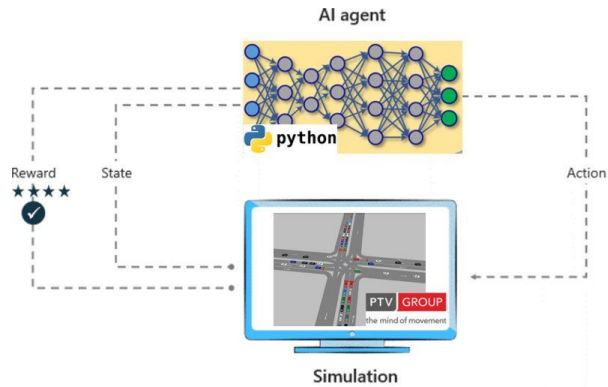

**Figure 2 DQN agent training process**

**Multiagent Deep Queue Networks**

Since this study considers each approaching vehicle to the intersection a separate agent, a proper DQN method capable of managing several agents must be deployed. The Multi-agent Deep Queue Network (MDQN) is designed to deal with sparse coordination settings in which an agent could be either in an independent or coordinated state. Being in an independent state does not require interactions with other agents, and the agent receives an individual reward (Equation (2)). Meanwhile, a coordinated state requires interaction with other agents, and the agent earns a joint reward (Equation (3)). The optimization goal is to maximize the total distributed reward at each step, known as the global reward (Equation (4)).

$$R(s_I, a_i) = r_i \tag{2}$$

$$r(s_j, a_j) = \sum_{b=1}^{h} r_{jb} \tag{3}$$

$$R(S, A) = \sum_{b=1}^{N} r_{(i \text{ or } j)\, b} \tag{4}$$

Where:

- Subscripts $i$ refer to an agent being in an independent state,
- Subscripts $j$ refer to an agent being in a joint state,
- $h$ is the number of agents involved in a joint state,

- N is the total number of agents in an environment,
- R is the global reward.

Referring to RL's nature, which considers future states and rewards while updating the current state's Q-value, a transition from an independent state to a coordinated state or vice versa requires specific settings for updating Q-value in MADQN. Three possible transition types appear in **Figure 3**.

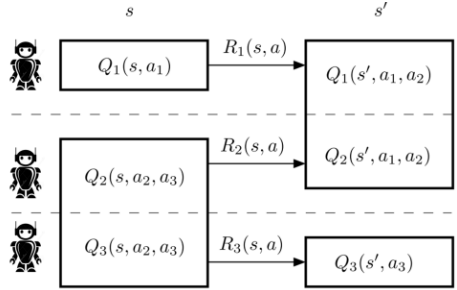

**Figure 3 Possible transition types in MADQN (24)**

Kok et al. (24) proposed formulations for updating Q-values for possible transition types in a conventional Q-learning approach. However, this study must adjust the formulas to be compatible with deep Q-learning. The adjustment includes removing the learning rate factor from the Q-values formulas since the learning rate is already included in neural network settings. The adjusted updating Q-values formulas for three possible types of transitions appear in the following.

Type 1: when an agent moves from a coordinated state to another coordinated state or from an independent state to another independent state, the coordinated or independent Q-value is updated by:

$$Q(s,a)_{i \text{ or } j} = r(s,a)_{i \text{ or } j} + \Upsilon \max Q(s',a')_{i \text{ or } j} \qquad (5)$$

Type 2: when an agent moves from a coordinated state to an independent state, the joint Q-value is updated by:

$$Q(s,a)_j = \sum_j^h [r(s,a)_j + \Upsilon \max Q(s',a')_i] \qquad (6)$$

Type 3: when an agent moves from an independent state to a coordinated state, the independent Q-values is updated by:

$$Q(s,a)_i = [r(s,a)_i + \Upsilon \frac{1}{h} \max Q(s',a')_i ] \qquad (7)$$

Based on the existing literature (7) (6), the pixel reservation system guarantees a collision-free vehicle maneuver system by itself. Therefore, unlike most other RL-based AV management systems, collision avoidance is not considered in the reward function as an optimization goal.

**EXPERIMENT DEVELOPMENT**

**Environment Design, States, and Reward Function**

The environment is a 4-way intersection including three lanes at each approach with dedicated left and right turn lanes. Traffic volume production, roadway configurations, and vehicles are set up in VISSIM software. The DSCLS application is developed in Python and coupled with VISSIM through the COM interface. The intersection area is divided into $2.5 \times 2.5$ m grid tiles, known as pixels or cells. The pixel reservation approach was initially proposed by Dresner et al. (6), and several researchers have used this method to manage CAVs at the intersection. This paper is inspired by a more recent pixel reservation-based study by Wu et al. (21).

The maximum speed is 40.2 km/hr, and the simulation resolution is set to 1 sec/step, meaning that each vehicle would be surveying a maximum length of 11.17 meters at each step ($V\Delta t = 11.7 \times 1 = 11.17$). Therefore, the grid zones are extended to 11.17 meters (5 cells) upstream of the intersection for each approaching section. The proposed algorithm takes control of the vehicles as soon as they enter the gridded area. Assuming a vehicle length of 4.90 meters (Ford Fusion 2019), each vehicle would occupy either two or three cells at each time step. The environment's physical settings and definition of current cells, desired cells, and commonly desired cells appear in **Figure 4**. The maximum value between survey and stop distances defines the number of desired cells for each vehicle. The survey distance includes the number of desired cells if the vehicle accelerates or keeps moving at the maximum speed. The stop distance covers the required cells for the vehicle to stop. Since, while reserving cells, it is unknown whether the vehicle would be commanded to accelerate or decelerate for the next step, it is required to reserve enough cells to cover both scenarios.

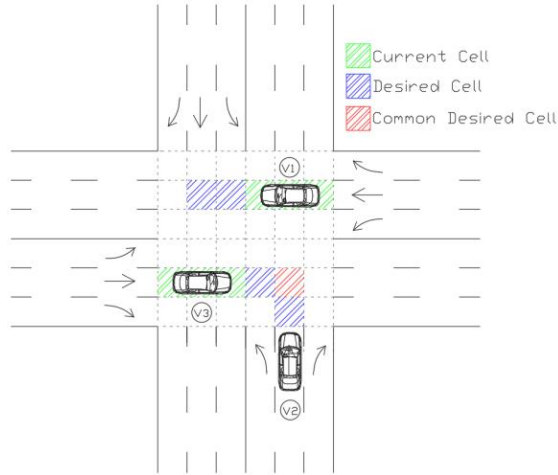

**Figure 4 The intersection environment physical settings**

Based on classical physics theory, the survey distance equals ($V\Delta t + \frac{1}{2}a\Delta t^2$ ) if the vehicle is moving slower than the maximum speed and can still accelerate, or it equals ($V\Delta t$) if the vehicle is already moving with the maximum speed, and the stop distance equals ($\frac{V^2}{2a}$). The acceleration and deceleration rates are set to 3.5 and 7 $m/sec^2$. The stop and survey distance formulas are shown in Equation (8).

$$DS = \begin{cases} \max\left(\left[\frac{V^2}{2a}\right], \left[V\Delta t + \frac{1}{2}a\Delta t^2\right]\right) & V < V_m \\ \max\left(\left[\frac{V^2}{2a}\right], [V\Delta t]\right) & V = V_m \end{cases} \quad (8)$$

Where:

- $\Delta t$: each step's time length
- V: current speed
- $a$: acceleration or deceleration capability
- DS: Number of desired cells

If the algorithm detects a shared desired cell between vehicles, they will enter a coordinated state; otherwise, they will be in an independent state. For example, vehicles 2 and 3 in **Figure 4** are in a coordinated state since they have reserved common cells. Each leading vehicle's state includes its current speed, current cell, and the queue length behind it. Possible actions for each vehicle to avoid collision are acceleration, deceleration, or maintaining the current speed. Equations (9) and Equation (10) present the state and possible actions.

$$State_i = (Speed_i, Current\ Cell_i, Queue_i) \quad (9)$$

$$Action_i = (-acc, +acc, maintain\ curent\ speed) \quad (10)$$

Each leading vehicle's reward at each time step is the summation of the delay of all vehicles in its direction. Assuming that the distance traveled by the vehicle $V_i$ at the time step t is $L_i$, then the optimal time required to pass the distance equals($\frac{L_i}{V_m}$). Each approach's delay value equals the reward function and is calculated based on Equation (11).

$$r(t) = -\sum_{i=1}^{n}(\Delta t - \frac{L_i}{V_m})] \quad (11)$$

The RL's goal is to optimize the total return (reward) at the end of each episode and not the immediate reward. Therefore, being independent in this environment does not necessarily result in acceleration or maintaining the maximum speed.

**Training Settings and Learning Performance**

For training purposes, traffic volume is set to 2,500 and 1,500 Veh/hr on major and minor streets. Vehicles are stochastically generated with a constant random seed by VISSIM software for 900 seconds. Initial Epsilon (ε) value is set to 1, which may cause long queues. Therefore, intersection legs are extended 2 km in each direction to reflect the potential queue length. The total epoch length is set to 1500 seconds to ensure that all generated vehicles have evacuated the intersection at the end of each epoch.

The simulation resolution is set to 1 step/sec, and the random seed is kept constant during all training iterations. As mentioned in the methodology section, to improve the DQN performance, a lagged copy of the neural network estimates the target Q-values. The target neural network parameters are updated every five epochs (target network update frequency). The replay memory size, which defines how many of the latest experiences will be stored in the memory for training purpose, is set to 10,000. The minibatch size, which defines the size of random data set from replay memory to be forwarded to the neural network, is set to 32. The training process hyperparameter settings appear in **TABLE 1**.

**TABLE 1 Training Process Hyperparameters Setting**

| Hyperparameter | Value |
| --- | --- |
| Number of steps per epoch | 1,500 |
| Step size (Δt), equal to simulation resolution | 1 sec |
| Replay memory size | 10,000 |
| Mini-batch size | 32 |
| Target network update frequency | 5 epochs |
| Initial ε | 1 |
| ε-decay factor | 0.999 |
| Discount (γ) | 0.9 |

**Figure 5** shows that the global reward converges to a maximum value after 3,500 training epochs, which means an overall delay reduction for the intersection.

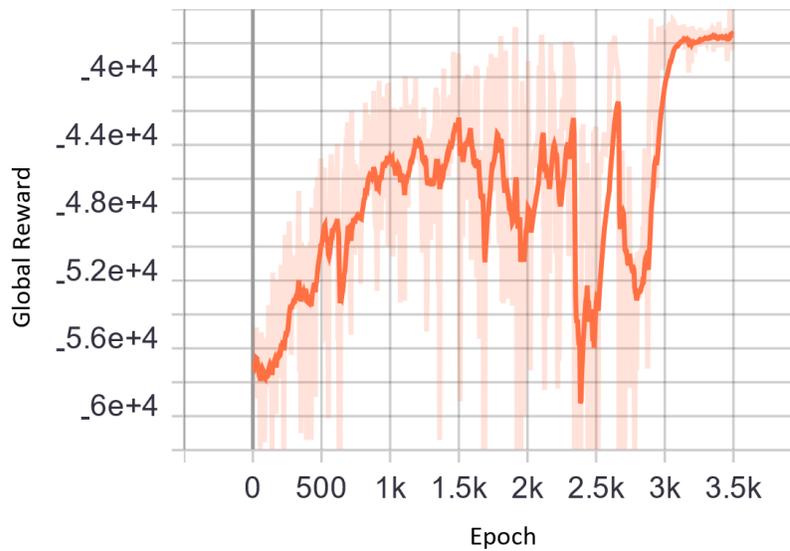

**Figure 5 Global reward convergence over training epochs**

**PROOF OF CONCEPT TEST**

The trained model is applied to a corridor consisting of four intersections to assess its impact on traffic flow in an extensive network. The test bed layout is shown in Figure 6.

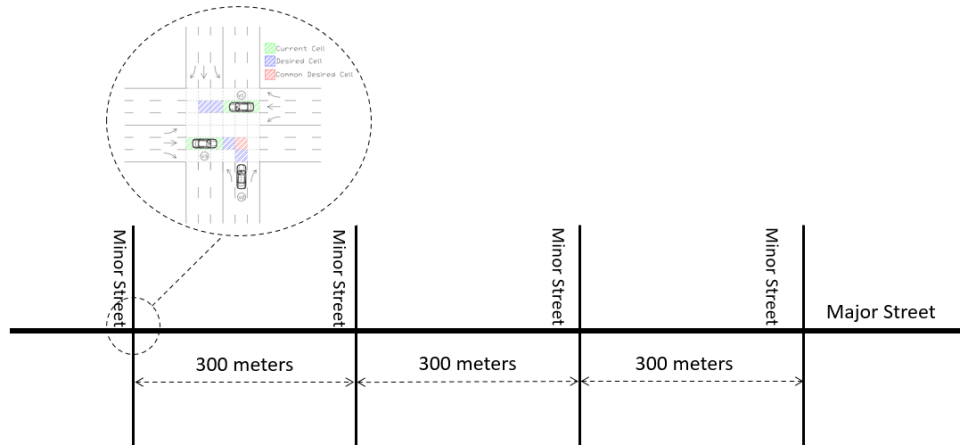

**Figure 6 Testbed layout**

The DSCLS performance is compared with other conventional and CAV-based intersection control systems, including 1) fixed traffic signal, 2) actuated traffic signal, and 3) Longest Queue First (LQF) control logic, which was developed by Wunderlich et al. (25). In this algorithm, the phases have no particular order and are triggered based on queue length in different approaches. In our study, the LQF logic is modeled based on pixel reservation for connected and autonomous vehicles, so the only difference between LQF and DSCLS would be the optimization approach. The speed limit is 40.2 km/hr, and three different volume regimes described below are considered for evaluation purposes:

1) Moderate volume: consists of 1,150 and 850 Veh/hr on the major and minor streets, approximately. This volume combination leads to Level Of Service (LOS) B for the major street and LOS C for the minor street.
2) High volume: consists of 1,600 and 1,100 Veh/hr on the major and minor streets, approximately. This volume combination leads to LOS D for both major and minor streets.
3) Extreme volume: consists of 2,000 and 1,300 Veh/hr on the major and minor streets, approximately. This volume combination leads to LOS F or congestion for major and minor streets.

The LOS is calculated based on the Highway Capacity Manual (HCM) 6$^{th}$ edition in Vistro, assuming the intersection operates under an optimized traffic light. Vistro is a traffic analysis and signal optimization software. Simulation time and resolution are set to 60 minutes and two steps/sec accordingly; 20 simulation runs with different random seeds are run in VISSIM software for each scenario.

**The Target Traffic Measure Comparison**

Since the control system's optimization goal is to minimize the delay, the delay is specified as the target traffic measure. A comparison of the average delay between different control systems appears in **Figure 7.** According to the figure, the DSCLS effectively reduces delay in all volume regimes, specifically in moderate volumes, with a 50% delay reduction compared to the second-best control system. A delay reduction of 29% and 23% is achieved compared to the second-best control systems in high and extreme volume regimes.

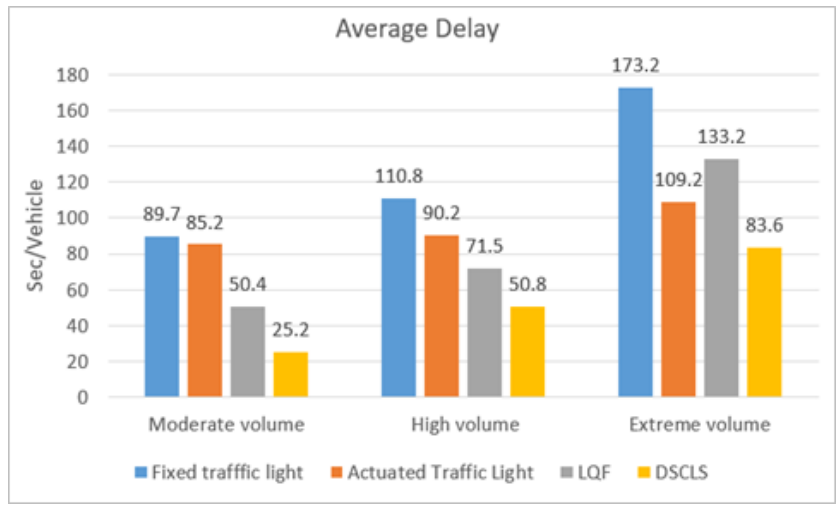

**Figure 7 Average delay comparison**

After delay reductions, travel time improvement is also expected. The average travel time values shown in **Figure 8**, reveal that 22% and 16% travel time reductions are gained compared to the LQF control system in moderate and high-volume circumstances. In the extreme volume regime, the proposed control system outperforms the actuated control system with an 11% travel time reduction.

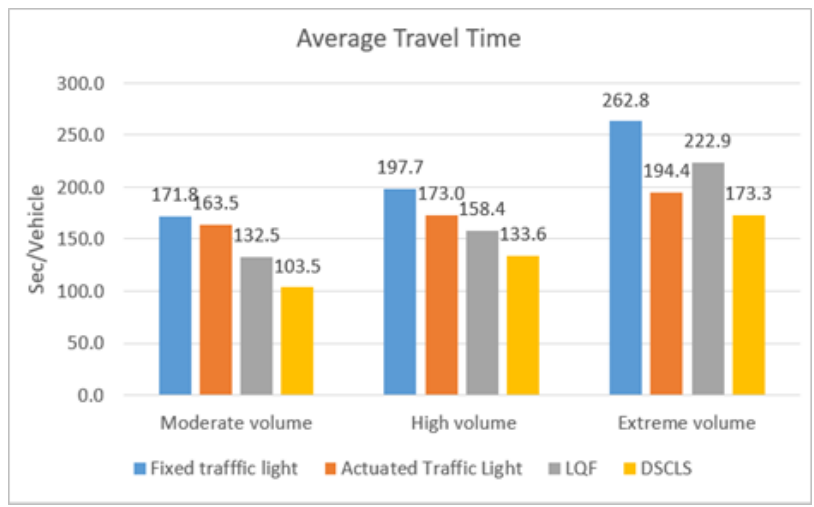

**Figure 8 Average travel time comparison**

**Other Measures Comparison**

Along with target traffic measures, the environmental and safety impacts of the model are also evaluated. Fuel consumption and $CO_2$ emission are calculated based on the VT-Micro model. This model can estimate emission and fuel consumption for individual vehicles based on instantaneous acceleration and speed (26). The average fuel consumption results appear in **Figure 9**; the DSCLS gains a 9% fuel consumption reduction in the moderate volume regime compared to the actuated control system. The actuated traffic light outperforms DSCLS with 2% and 6% fuel consumption reductions in high and extreme volume regimes. However, the proposed control system noticeably outperforms the other pixel reservation-based model (LQF), in all volume regimes.

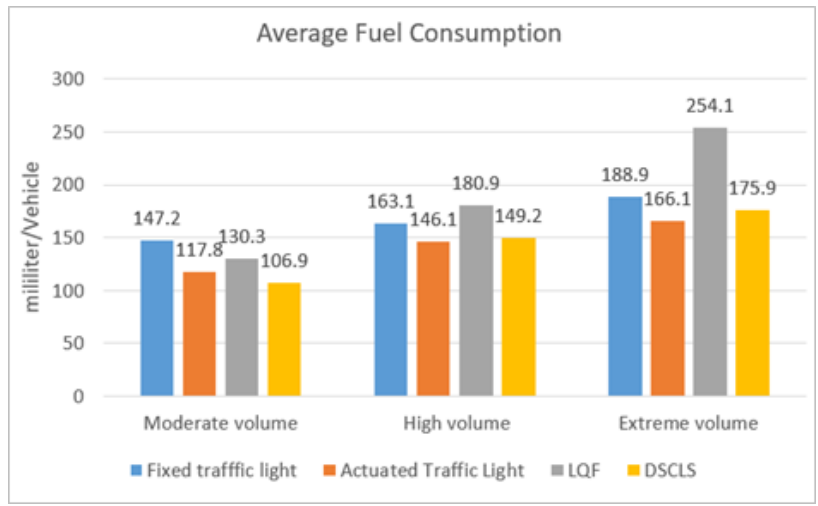

**Figure 9 Average fuel consumption**

As shown in **Figure 10**, average $CO_2$ emission follows the same pattern as average fuel consumption. The DSCLS outperforms actuated traffic lights with a 9% $CO_2$ emission improvement. However, the actuated traffic light performs better than DSCLS with 4% and 7% less CO2 emission in high and extreme volume regimes.

Despite acceleration and deceleration rates for DSCLS being limited to fixed values only, fuel consumption and emission improve slightly in moderate and high-volume regimes, and a slight increment occurs in the extreme volume condition.

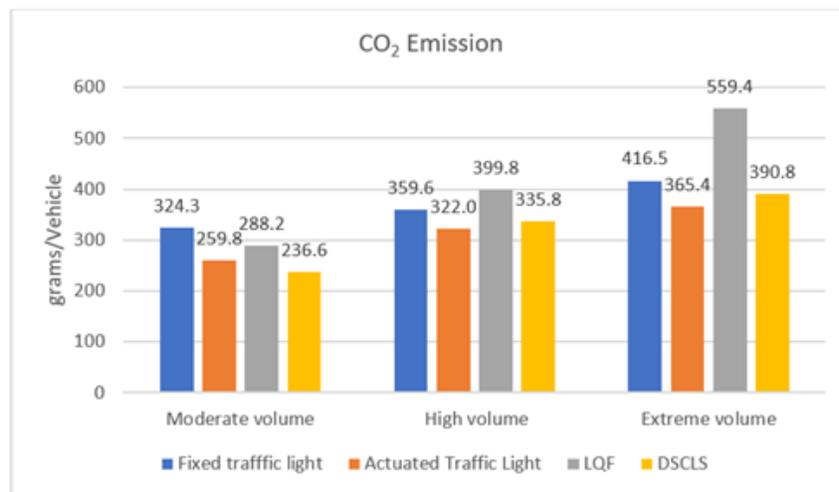

**Figure 10 Average $CO_2$ emission**

Among various Surrogate Safety Measures, the Post Encroachment Time (PET) is a well-fitting measure to identify safety threats for crossing vehicles at an intersection. It represents the time between the departure of the encroaching vehicle from the conflict point and the vehicle's arrival with the right-of-way at the conflict point (27). Safety analysis is performed in the SSAM software, automatically identifying, classifying, and evaluating traffic conflicts in the vehicle trajectory data output from microscopic traffic simulation models. The comparison of PET for different traffic control systems appears in **Figure 11**.

According to the figure, pixel reservation-based logics, including DSCLS and LQF, have almost equal PET. Both perform better than the actuated traffic light, specifically in extreme volume regimes with a 22% improvement in PET. Pixel-reservation-based control systems consist of more stop-and-go, which increases the chance of an accident in regular intersections. Therefore, even minor improvement in PET compared to conventional control systems is noticeable.

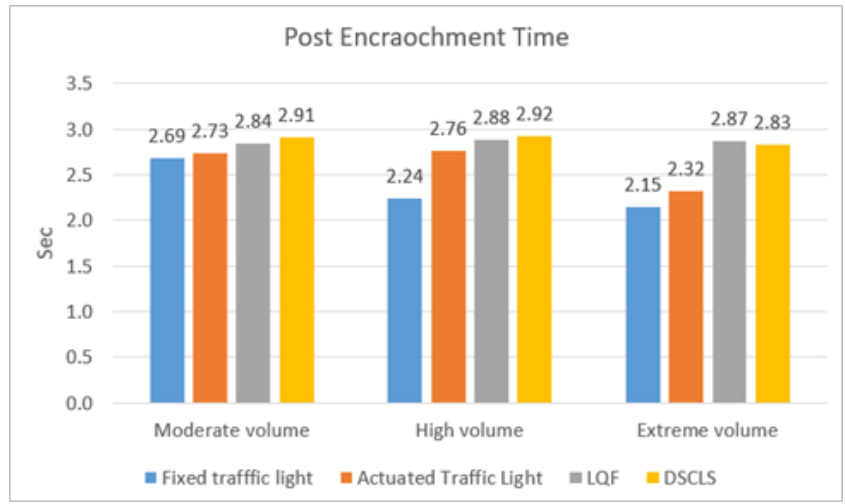

**Figure 11 Post encroachment time comparison**

**Statistical Analysis of the Simulation Results**

According to the t-test results performed on the simulation results between the DSCLS and other control systems, gains from DSCLS compared to the fixed traffic light are statistically significant for all measures in all volume regimes. The proposed model's minor losses or gains in fuel consumption and emission are not statistically significant compared to the actuated traffic lights. DSCLS gains compared to LQF are statistically significant in all measures except for safety, which is predictable since both models are based on pixel reservation logic to avoid collisions. The t-test results appear in **TABLE 2**.

**TABLE 2 P-Values For T-Test Results**

| | Delay | Travel Time | Fuel Consumption | $CO_2$ Emission | PET |
|---|---|---|---|---|---|
| **Fixed Traffic Light** | | | | | |
| *Moderate volume* | | | | | |
| | <0.05 | <0.05 | <0.05 | <0.05 | <0.05 |
| *High Volume* | | | | | |
| Delay | Travel Time | Fuel Consumption | $CO_2$ Emission | PET | |
| <0.05 | <0.05 | <0.05 | <0.05 | <0.05 | |
| *Extreme volume* | | | | | |
| Delay | Travel Time | Fuel Consumption | $CO_2$ Emission | PET | |
| <0.05 | <0.05 | <0.05 | <0.05 | <0.05 | |
| **Actuated Traffic Light** | | | | | |
| *Moderate volume* | | | | | |
| Delay | Travel Time | Fuel Consumption | $CO_2$ Emission | PET | |
| <0.05 | <0.05 | >0.05 | >0.05 | <0.05 | |
| *High Volume* | | | | | |
| Delay | Travel Time | Fuel Consumption | $CO_2$ Emission | PET | |
| <0.05 | <0.05 | >0.05 | >0.05 | <0.05 | |
| *Extreme volume* | | | | | |
| Delay | Travel Time | Fuel Consumption | $CO_2$ Emission | PET | |
| <0.05 | <0.05 | >0.05 | >0.05 | <0.05 | |
| **LQF** | | | | | |
| *Moderate volume* | | | | | |
| Delay | Travel Time | Fuel Consumption | $CO_2$ Emission | PET | |
| <0.05 | <0.05 | <0.05 | <0.05 | >0.05 | |
| *High Volume* | | | | | |
| Delay | Travel Time | Fuel Consumption | $CO_2$ Emission | PET | |
| <0.05 | <0.05 | <0.05 | <0.05 | >0.05 | |
| *Extreme volume* | | | | | |
| Delay | Travel Time | Fuel Consumption | $CO_2$ Emission | PET | |
| <0.05 | <0.05 | <0.05 | <0.05 | >0.05 | |

(The three main sections — Fixed Traffic Light, Actuated Traffic Light, and LQF — are each labeled **DSCLS** along the left margin.)

## CONCLUSIONS

This paper presented a signal-free intersection control system for CAVs. The control system is based on a pixel reservation logic to detect potential colliding movement and a DSCLS, with the optimization goal of minimizing the overall delay at the intersection. The traffic volume and vehicle arrival's random seed are constant during 3,500 epochs of model training iterations. However, the trained model can deal with any volume combination or stochastic vehicle arrivals due to the chain impact of the agent's random actions in a DRL training course, known as exploration. The DSCLS is developed as a Python application, controlling individual vehicles in VISSIM software.

Performance of the proposed model is compared with conventional and CAV-based intersection control systems, including fixed traffic lights, actuated traffic lights, and LQF, in a corridor of four intersections. Based on the simulation results, the DSCLS noticeably outperforms the LQF control system with up to 50% and 20% dealy and travel time reduction. Both CAV-based control systems have shown the same performance in safety measures. However, the DSCLS outperforms the LQF in fuel consumption and $CO_2$ emission by up to 30%. Although the proposed model is based on fixed acceleration and deceleration rates, it does not cause any significant losses in fuel consumption and $CO_2$ emission compared to the conventional intersection control systems, which is a promising result.

## CONTRIBUTIONS AND FUTURE DIRECTIONS

Based on the existing literature, deep reinforcement learning is an outstanding decision-making logic in stochastic environments. To the best of the author's knowledge, this research is the first try at deploying deep reinforcement learning to manage CAVs' conflicting maneuvers in a roadway network setting. The developed CAV management system can improve its performance or adjust to the new circumstances during its life cycle, which is the main challenge in conventional optimization-based control systems. The proposed algorithm assumes individual vehicles as independent agents, each having its own optimization logic, and coordinating with other agents only if required, which is most likely how highly automated vehicles will act in the future.

This study's future directions can be 1) developing a passenger throughput optimization-based or emergency vehicle priority-based control system by adjusting the reward function, 2) assuming several accelerations and declaration rates for each vehicle to improve environmental measures, and 3) using other DRL methods such as Double DQN and Prioritized Experience Replay to improve DQN's performance.